\documentclass[11pt,a4paper]{article}
\usepackage[hyperref]{iwpt2021}
\usepackage{url}
\usepackage{latexsym}
\usepackage{graphicx,dblfloatfix}
\usepackage{hyperref}
\usepackage{times}
\usepackage{xcolor}
\usepackage{latexsym}
\usepackage{csquotes}
\usepackage{multicol,array,booktabs,siunitx,multirow,dcolumn}
\usepackage{microtype}
\usepackage[title]{appendix}
\usepackage{tikz-dependency,subcaption,float}
\usepackage{amsmath,multirow}
\usepackage{amssymb,amsmath}
\usepackage{nicefrac}
\aclfinalcopy 

\definecolor{deppink}{HTML}{a92b7a}
\definecolor{depgreen}{HTML}{679033}
\definecolor{depgreen2}{HTML}{74b038}
\definecolor{deporange}{HTML}{be6320}
\definecolor{depblue}{HTML}{165f77}
\definecolor{depred}{HTML}{b12019}
\definecolor{depgrey}{HTML}{6c8093}
\definecolor{white}{HTML}{ffffff}
\definecolor{deppurple}{HTML}{4888ab}
\definecolor{deppurple2}{HTML}{565393}

\newcommand\MyLBrace[2]{%
#2$\left.\rule{0pt}{#1}\right\{$}

\makeatletter
\renewcommand{\boxed}[1]{\text{\fboxsep=.2em\fbox{\m@th$\displaystyle#1$}}}
\makeatother
  
\title{A Modest Pareto Optimisation Analysis of Dependency Parsers in 2021}

\author{Mark Anderson \\
 Universidade da Coru\~na, CITIC\\
 Department of CS \& IT \\
  {\tt m.anderson@udc} \\\And
    Carlos G{\'o}mez-Rodr{\'\i}guez \\
 Universidade da Coru\~na, CITIC\\
 Department of CS \& IT \\
  {\tt carlos.gomez@udc.es} \\}

\date{}

\begin{document}
\maketitle
\begin{abstract}
  We evaluate three leading dependency parser systems from  different paradigms on a small yet diverse subset of languages
  in terms of their accuracy-efficiency Pareto front. As we are interested in efficiency, we evaluate core parsers without pretrained language models (as these are typically huge networks and would constitute most of the compute time) or other augmentations that can be transversally applied to any of them. Biaffine parsing emerges as a well-balanced default choice, with sequence-labelling parsing being preferable if inference speed (but not training energy cost) is the priority.
\end{abstract}

\section{Introduction}
The inefficiency of modern NLP systems has recently come under scrutiny, especially regarding their large energy consumption \cite{strubell2019energy}. This hasn't started a revolution, but there is some NLP work where efficiency is considered. \citet{zhang-duh-2020-reproducible} studied different 
settings
for neural machine translation systems, evaluating not only accuracy but also certain costs such as inference time, training time, and model size. \citet{zhou-etal-2021-hulk} analysed the fine-tuning and inference time for pretrained LMs, and estimated the cost of pre-training. \citet{jacobsen2021optimal} presented a Pareto optimisation analysis for POS taggers, considering accuracy and model size. 

In parsing in particular, \citet{strzyz2019viable} evaluated dependency parsing as sequence labelling specifically to increase inference efficiency and also undertook a Pareto optimisation analysis. Others used model compression via distillation to increase inference speed of neural parsers with a mixed bag of results \cite{dehouck-etal-2020-efficient,anderson-gomez-rodriguez-2020-distilling}.
\citet{dehouck-etal-2020-efficient} also took into consideration the training energy costs of distilling models, which highlighted the high energy cost of this technique. 

We present a Pareto optimisation analysis on modern dependency parsing systems. We cover three systems which are broadly representative of current approaches. We analyse their efficiency with respect to inference speed and also their training cost, measured in energy consumption.

\paragraph{Contribution:} A simple, modest analysis on the merits of different parser systems that cover three current paradigms. 
Our goal is not to provide surprising results, but a realistic snapshot of the current state of affairs of a representative sample of modern parsing systems on linguistically diverse data. 
This analysis runs the systems in a consistent way with respect to software, hardware, and network settings. We also offer a brief overview of self-reported performance on PTB for systems that have a published speed. We add to this measurements for a subset of these systems which we ran locally for a more consistent comparison, 
i.e.\ something of a reproducibility effort.
\paragraph{Disclaimer} We make a practical comparison for practitioners, so we focus on publicly available systems on typical hardware that doesn't require a huge budget. We are not making general claims that technique X is always more efficient than technique Y in the abstract or that this will hold in any hardware. Also, the extent to which an implementation has been engineered will impact performance, so we have referenced the original repositories used.\footnote{The moderately edited code is available at \url{http://www.grupolys.org/software/iwpt2021/parsers-code.zip}.} 
\section{PTB performance}

\begin{table*}[htpb!]
\small
    \centering
    \begin{tabular}{lcccc}
    \toprule
     \multicolumn{1}{c}{} & \multicolumn{2}{c}{\textbf{speed (sent/s)}} \\
       \multicolumn{1}{c}{}  & \multicolumn{1}{c}{\textbf{GPU}} & \multicolumn{1}{c}{\textbf{CPU}} & \textbf{UAS} & \textbf{LAS} \\\midrule
         HPSG \citep{zhou2019head} & 159$^{\ast}$ & - & 96.09$^{\ast}$ & 94.68$^{\ast}$ \\
         Biaffine w\ CRF \cite{zhang-etal-2020-efficient} & 400$^\ast$ & & 96.14$^\ast$  &  94.49$^\ast$\\
       Pointer-LR \citep{fernandez2019left}& 23$^{\ast}$  & - & 96.04$^{\ast}$ & 94.43$^{\ast}$ \\
    
    GNN \cite{ji-etal-2019-graph} & 416$^\ast$ & - & 95.97$^\ast$ &   94.31$^\ast$\\
        Pointer-TD \citep{ma2018stack}& 10.2$^{\dagger}$ & - & 95.87$^{\ast}$ & 94.19$^{\ast}$ \\
    Biaffine \citep{dozat20161} & 411$^{\ast}$ & - &95.74$^{\ast}$ &  94.08$^{\ast}$\\ Distilled-Ensemble \cite{kuncoro-etal-2016-distilling} & - & 20$^\ast$ & 94.26$^\ast$ & 92.06$^\ast$ \\
    BIST - Transition \citep{kiperwasser2016simple} & - & 76$\pm$1$^{\ddagger}$ & 93.9$^{\ast}$ & 91.9$^{\ast}$\\
    SeqLab \citep{strzyz2019viable} & 648$\pm$20$^{\ast}$ & 101$\pm$2$^{\ast}$ & 93.67$^{\ast}$ & 91.72$^{\ast}$   \\
    BIST - Graph \citep{kiperwasser2016simple}& - & 80$\pm$0$^{\ddagger}$ & 93.1$^{\ast}$ & 91.0$^{\ast}$ \\
    
    CM \citep{chen2014}& - & 654$^{\ast}$ & 91.80$^{\ast}$ & 89.60$^{\ast}$  \\
    \midrule
    Pointer-LR & 95$\pm$1  & 8$\pm$0 &  96.02 & 94.47\\
    Biaffine (PyTorch) & 1003$\pm$3 & 53$\pm$0 & 95.74 & 94.07 \\
    UUParser \citep{smith201882} &-& 42$\pm$1& 94.63 & 92.77 \\
    Distilled-Biaffine \cite{anderson-gomez-rodriguez-2020-distilling} & 1153$\pm$3 & 96$\pm$0 & 94.59 & 92.64 \\
    SeqLab  & 1064$\pm$13 & 99$\pm$1 & 93.46 & 91.49 \\
    MaltParser 1.9.2 w/ Stack lazy \cite{nivre2007maltparser} & - & 473$\pm$11 & 89.29 & 86.95 \\
     \bottomrule
    \end{tabular}
    \caption{
    Performance for current leading parsers for the English PTB with POS tags predicted from the Stanford POS tagger.  $\ast$ denotes values taken from the original paper, $\dagger$ from \citet{fernandez2019left}, and  $\ddagger$ from \citet{strzyz2019viable}. Values with no superscript 
    are from running the models on our system locally 
    (speeds averaged over 5 runs) and with a batch size of 256 (excluding UUParser which doesn't support batching) with GloVe 100 dimension embeddings. 
    Table is extended from one 
    in \citet{anderson-gomez-rodriguez-2020-distilling}.
    }
    \label{tab:current-state}
\end{table*}

For historical reasons, it is common practice for parsers to report performance results on the English Penn Treebank (PTB) \cite{marcus1993building}. While such results at best provide a partial picture on a single language, they are by far the most comprehensive source of results provided in the literature under a consistent context (at least in terms of data and splits, although not hardware), so they are useful to see high-level trends and as a starting point to choose parsers for our experiment.


In Table \ref{tab:current-state} we report performance of modern parsing systems for which speeds have been reported. We couldn't find a reported speed of \citet{clark-etal-2018-semi} which currently has the highest reported performance on PTB (UAS 96.61 and LAS 95.02) when not using BERT.
However, its main contribution is semi-supervised augmentations that could be utilised by any parsing system, with their core parser being the Biaffine parser.
\citet{zhou2019head}'s
system leverages
constituency and dependency parsing and when 
not using training data with both constituency and dependency annotations (often not available)
the system achieves UAS 95.82 LAS 94.43 (i.e.\ very similar in LAS to the other top-performing sytems). 
\citet{zhang-etal-2020-efficient} use a Biaffine parser but with a moderate beam search, which is obviously less efficient than the original. It results in a small increase in performance. \citet{ji-etal-2019-graph} use graph neural networks to learn enriched high-order information from partial parses. It again only gains small increases over Biaffine, but is more computationally complex 
and code is not available.

We report results for UUParser of \citet{smith201882} that we ran locally 
(refreshingly the original paper didn't use PTB).
While the results show a reasonable speed-accuracy trade-off, we
opted not to use this for the current analysis as the original code is implemented in DyNet which doesn't properly support CUDA,
and is a different framework from that of the other parsers we opted to choose.

Based on this, we opted to use the basic Biaffine parser to represent graph-based parsers, the Pointer-LR network as the representative of transition-based algorithms,\footnote{Some might argue that it isn't a clear cut case of a transition-based parser, but it transitions from state to state like more traditonal algorithms.} and the sequence-labelling parser 
to represent SL systems.
They all have the added benefit of working under the same software and having code available. 

Note that, as we make emphasis on efficiency, we focus on reasonably bare-bones versions of the parsers. The impact of pretrained language models, or other augmentations that are transversal to the parsing system, is outside the scope of this paper.

\section{Pareto optimisation analysis}
Here we detail the parsing systems, the data we used, and how model structures were altered.

\subsection{Parsers}
All the parsers use BiLSTMs, but have additional structures which set them apart from one another and use one of three paradigms broadly speaking: one is a transition-based parser, one is a sequence-labelling parser, and the last is a graph-based parser. For space reasons, we only very briefly outline them here, but give more details in Appendix \ref{app:parsers}.
\paragraph{Left-to-right pointer network} (L2R). One of the current top-performing parsers on PTB, it uses a left-to-right transition-based algorithm that builds a number of attachments equal to sentence length using a pointer network~\cite{ma2018stack,fernandez2019left}.\footnote{https://github.com/danifg/SyntacticPointer.} 

\paragraph{Deep biaffine} (\textsc{Biaffine}) \cite{dozat20161} is an edge-factored graph-based parser that produces a matrix of scores giving a probability distribution on arcs, where the Chu-Liu-Edmonds algorithm \cite{chuliu65,edmonds1967optimum} is then applied to obtain a tree.


\paragraph{Sequence labelling parser} (\textsc{SeqLab}) 
encodes trees as a sequence of labels, so that a direct one-to-one prediction can be made for each token in a sentence \cite{spoustov2010dependency,li-etal-2018-seq2seq,strzyz2019viable}.\footnote{We use refactored encoding/decoding functions from https://github.com/mstrise/dep2label.} 
We implement it using the Biaffine system described above (for uniformity) editing it to be a sequence-labelling system. 

\subsection{Data}

In our choice of treebanks, we balance three factors: the need to use a small number of treebanks (as our detailed Pareto analysis implies training a large number of models per treebank), linguistic diversity and treebank quality.
This leads us to choose 4 high-quality (manually annotated or corrected, and relatively large) treebanks covering 3 different language families and 4 subfamilies: UD-Hindi-HDTB, UD-Polish-PDB, UD-Korean-Kaist and the Chinese Penn Treebank. More details of each treebank, justifying their diversity and adequacy for the analysis are given in Appendix \ref{app:data}.


\begin{figure*}[tb!]
    \centering
    \includegraphics[width=0.9\linewidth]{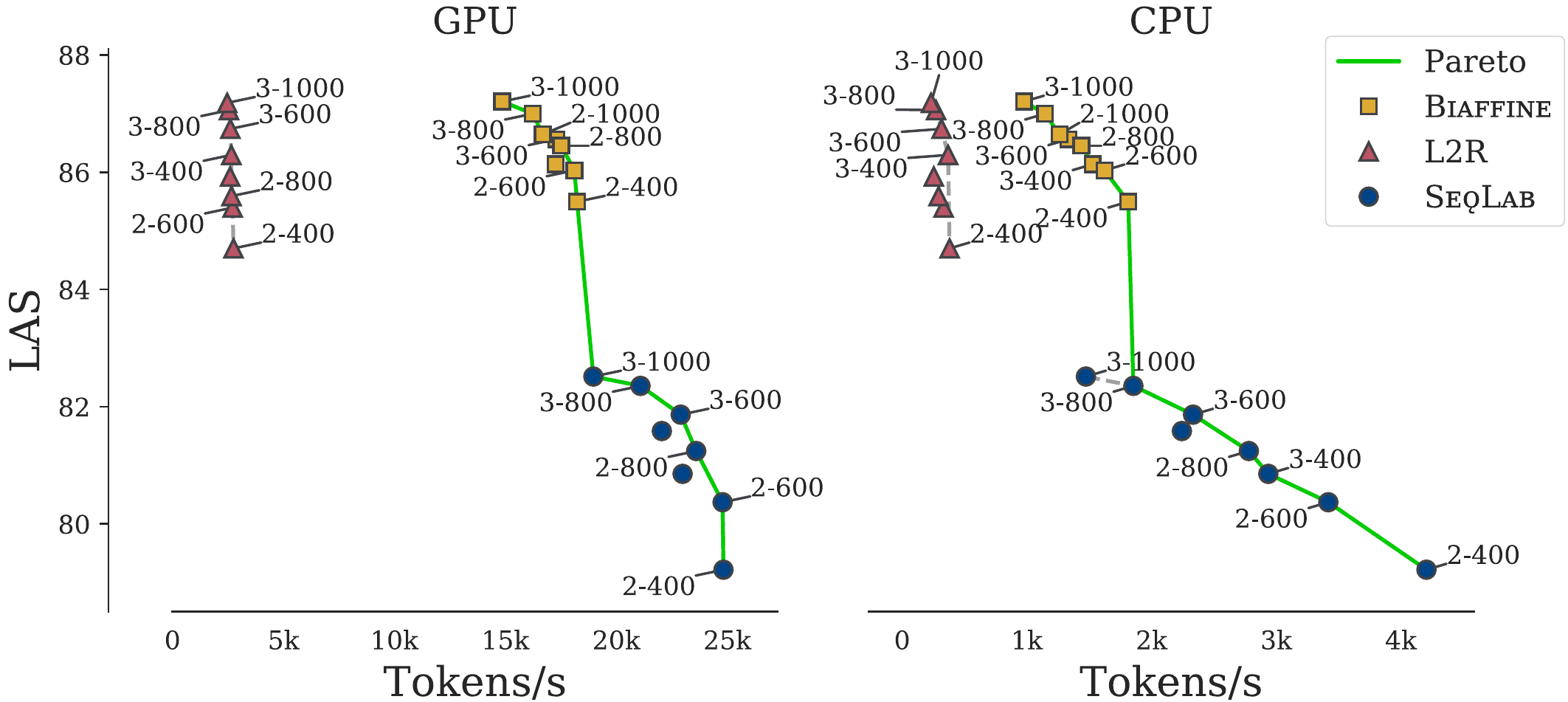}
    \caption{Pareto fronts for L2R, \textsc{Biaffine}, and \textsc{SeqLab} for the development data.}
    \label{fig:speed_paretos}
\end{figure*}
\subsection{Methodology}

We vary the size of the BiLSTM component of the networks by their number of layers and nodes. Each parser 
has randomly-initialised character embeddings and pretrained word embeddings as only inputs.
We use pretrained FastText embeddings \cite{grave2018learning}. Except for Chinese, as the FastText embeddings are in the traditional script, so we use the embeddings from \citet{shen_emb18}.\footnote{\url{https://jima.me/open/cwv/}} The embeddings are reduced to 100 dimensions using PCA. The structure of the networks are very similar. The L2R system uses a biaffine transformation to score the transitions at each step similar to the \textsc{Biaffine} parser, and we use the same sizes for the layers.
The \textsc{SeqLab} system is altered from the \textsc{Biaffine} implementation and is exactly the same except the layers needed for the biaffine transformation are replaced by two MLPs which predict the 
labels
for each token. The only major difference in the networks is that L2R uses a CNN to create the character embeddings and the other two use BiLSTMs. We didn't change this 
in order to avoid modifications to the systems.
The network hyperparameters are shown in Table \ref{tab:experimental_hyperparameters} in Appendix \ref{app:hyp}. Models were trained on GPU, but we report the energy used by both the GPU and CPU.

We could have altered other aspects of the network, but the main computational cost comes from the BiLSTM layer. The other main contender to alter would be the embedding layers. For example, we could have altered the size of the character BiLSTM/CNN, but certain experiments show that it has a limited impact on accuracy \cite{smith2018investigation,anderson-gomez-rodriguez-2020-frailty}.

We measured the speed of each system on each treebank by running them 5 times using a single CPU core, both for speeds measured running on GPU and CPU, so that we get a reasonably accurate measure of the speed for each treebank. We then report macro averaged speeds across treebanks.

We use the \texttt{energyusage} package for measuring training energy.\footnote{\url{https://pypi.org/project/energyusage}} It measures the power usage of the GPU and CPU while a process is running (having taken a measure of the background usage). We minimised the use of the system when training these models to obtain accurate measurements, but they aren't overly precise. This isn't a major issue as the measurements are over long periods of time and so unless there were massive fluctuations when training a given model, comparison is fine. We use joules (or kJ and MJ) as they are the SI units for energy \cite{sibrochure} and, unlike carbon emissions, they are independent of external factors like regional electricity generation grids.\\
\textbf{Hardware:} Intel Core i7-7700 and Nvidia GeForce GTX 1080. \\
\textbf{Software:} Python 3.7.0, PyTorch 1.0.0, and CUDA 8.0.

\subsection{Pareto fronts: inference speed}
Figure \ref{fig:speed_paretos} shows LAS versus parsing speed for the development data (we also present the same for the test data in Figure \ref{fig:speed_paretos_test} in the Appendix that echoes the visualisation seen here). The individual Pareto front for each parser is shown (light grey, dashed) 
As expected, models with larger networks are more accurate but slower.
More interestingly, the overall Pareto front
is exclusively constructed of \textsc{Biaffine} and \textsc{SeqLab} systems. While L2R does achieve similar accuracy scores as \textsc{Biaffine}, it is considerably slower. \textsc{SeqLab} is the fastest option by a clear margin (especially smaller networks on CPU). 
So the practical advice to draw from this aspect or the Pareto optimisation would be to use \textsc{Biaffine} if accuracy is the main concern, or \textsc{SeqLab} if inference time is important.

\subsection{Pareto fronts: training energy}
\begin{figure}[b!]
    \centering
    \includegraphics[width=0.9\linewidth]{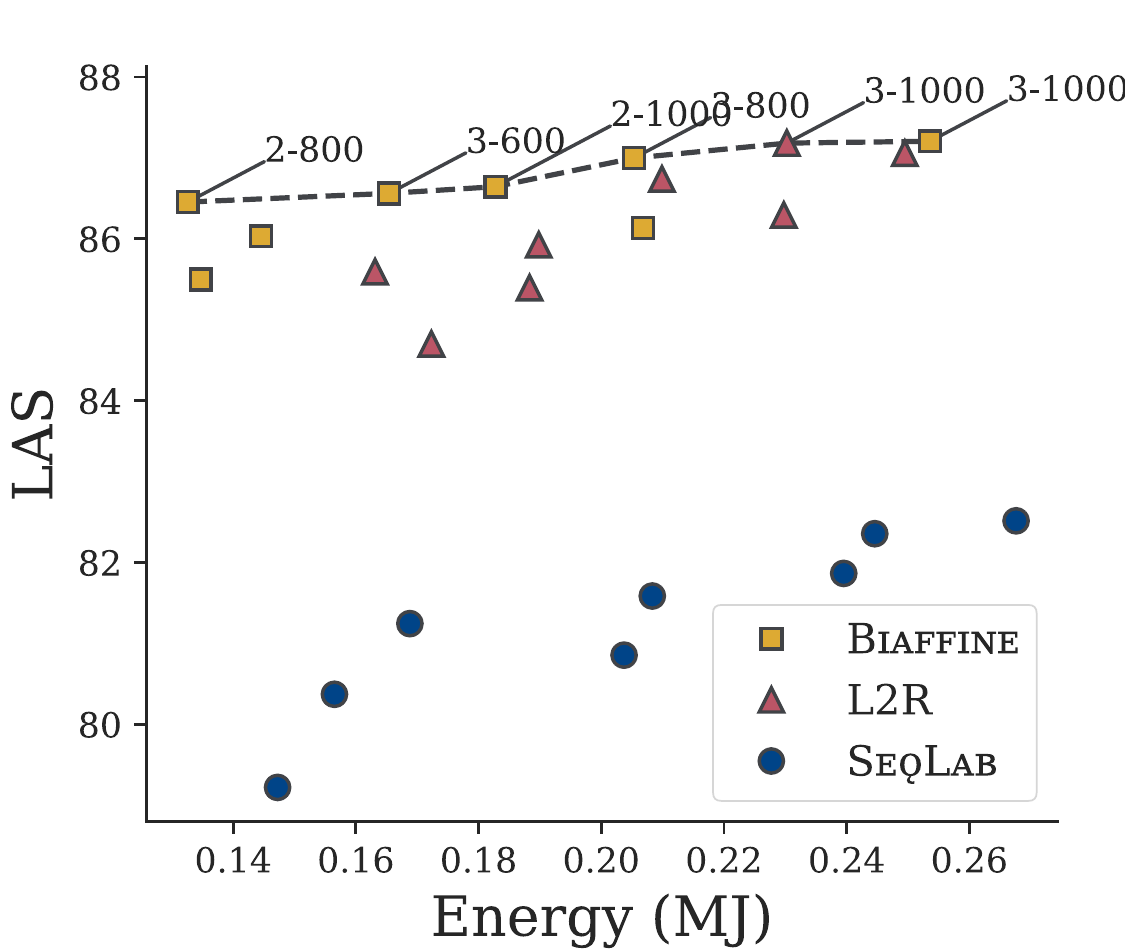}
    \caption{Pareto fronts for L2R, \textsc{Biaffine}, and \textsc{SeqLab} for training energy.}
    \label{fig:energy_paretos}
\end{figure}
Figure \ref{fig:energy_paretos} shows LAS against the average energy (across treebanks) consumed during training (in training, we always use the GPU). There is no clear link between the energy consumed and the accuracy of a system. However, this visualisation highlights that \textsc{SeqLab} is nowhere near optimal with respect to training efficiency.

\begin{figure}[b!]
    \centering
    \includegraphics[width=0.9\linewidth]{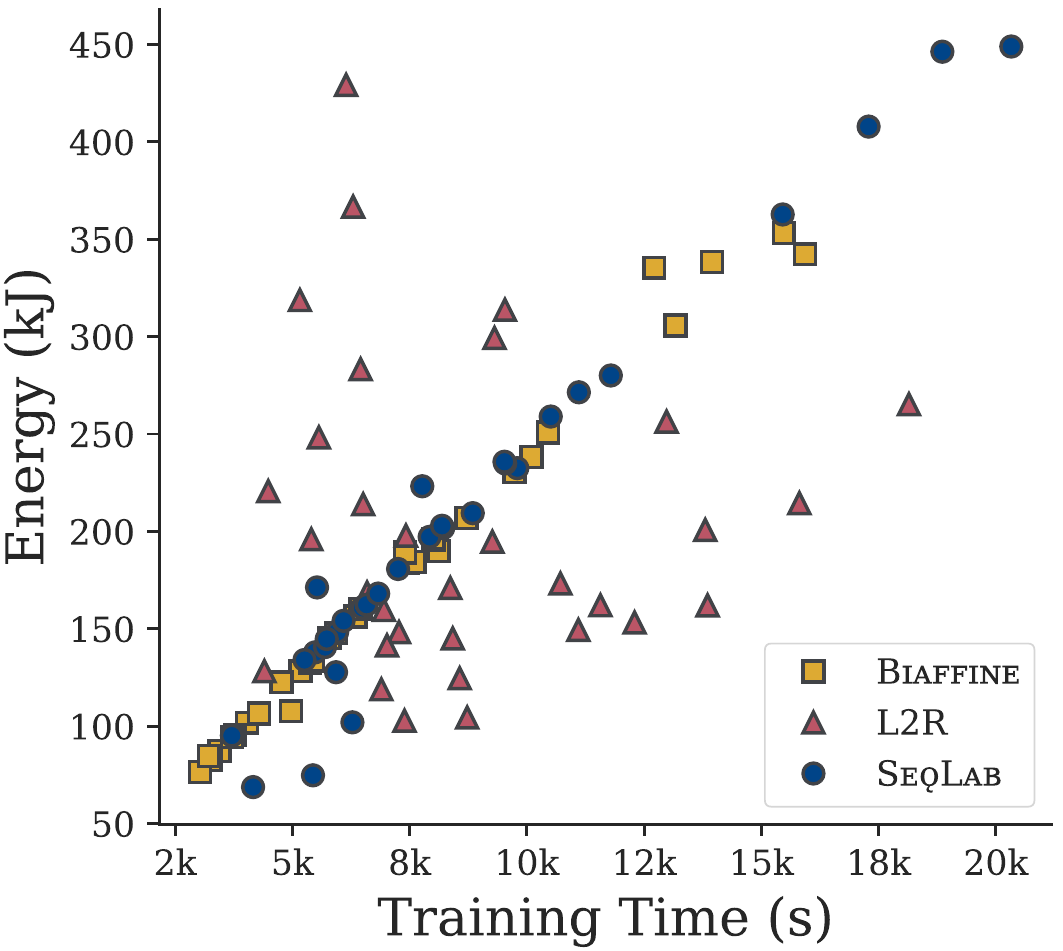}
    \caption{Training energy consumption with respect to training time.}
    \label{fig:energy_time}
\end{figure}

The amount of energy consumed during training is basically dependent on the time it takes each system to converge as can be seen in Figure \ref{fig:energy_time}. In this figure, we show 
individual models
(i.e.\ not averaged over treebanks). The relation for \textsc{Biaffine} and \textsc{SeqLab} is very clearly linear between energy and training time, suggesting that there is nothing intrinsically more energy consuming between these systems beyond convergence time. 
For L2R, this relation seems to hold broadly, but is less clear. It appears that L2R is more sensitive to the nature of the data, which we expand on in Appendix \ref{app:l2r_energy}.  
\section{Limitations of analysis}\label{app:limits}
While our analysis is not ground-breaking or particularly expansive in nature, we do think it is useful in practice
and acts as mini-review of the current state of affairs in dependency parsing. However, there are a number of limitations in this study. First, we only look at the parameters associated with the BiLSTMs. We feel this is fairly justified, but it is obviously feasible that varying these parameters and not the others could have different effects for each parsing system even if that is fairly unlikely. While we do look at a very diverse set of languages with diverse linguistic features, it is still a fairly small sample. We were somewhat limited by having to train many models and felt it would be better to focus on a sample of diverse languages with quality data than many languages and less model settings. Of course, this analysis could be extended to use more languages, but we expect this would further corroborate the results presented here. Also by using a small set of treebanks, we don't cover a wide array of domains (the data is mainly fiction and news).

Another potential limitation is only using one dependency annotation scheme (the scheme used for CTB was a precursor to UD), but in lieu of a theoretical reason that the parsers would behave differently using a different scheme (e.g.\ surface syntactic UD (SUD) treebanks containing much more non-projectivity \cite{gerdes-etal-2018-sud}) this feels like a light limitation. 

A slightly more pressing limitation is the absence of a feature analysis because certain systems could potentially benefit from different features. Work has been presented in this direction and has shown that predicted POS tags aren't wonderfully useful \cite{smith2018investigation,anderson-gomez-rodriguez-2020-frailty,zhang2020pos}. However, these analyses didn't include \textsc{SeqLab} parsers at all and the transition-based system used was 
a lower-performing system,
UUParser. So it is feasible that L2R and \textsc{SeqLab} would benefit from predicted POS tags. That can be left open for the future. 

Another limitation is that we only trained one model for each BiLSTM setting. While training a model for each treebank somewhat offset this, it is still possible that with different initialisation, these parsers would behave slightly differently. However, it is unlikely to cause material differences in the performance and as mentioned, this is quite strongly offset by training on varying treebanks. 

And finally, we focused on parsers trained on fairly large amounts of annotated data. We leave the analysis of different parsing systems in a low-resource setting for others, but we point out that when training on very little data, training costs aren't much of a concern and 
on truly low-resource languages, data parsed at production is also going to be scarce so inference speed won't be the bottleneck.
\section{Conclusion}
We have presented a simple Pareto optimisation analysis for a representative sample of modern dependency parsers. We evaluated efficiency in two ways. We evaluated the trade-off between accuracy and parsing speed and the trade-off between accuracy and training energy consumption. The \textsc{Biaffine} and \textsc{SeqLab} occupied the speed Pareto front with the former being slower and more accurate and the latter being faster and less accurate. We didn't observe any real trade-off with regards to training energy and performance, but it was clear that \textsc{SeqLab} is not particularly efficient in this regard. Typically training energy varied based on how long a model took to converge, with L2R being somewhat sensitive to the different treebanks. 
Overall, for most scenarios, \textsc{Biaffine} emerged as a well-balanced practical solution. For the sake of candour, we offer a brief discussion of the limitations of this analysis in Appendix \ref{app:limits}.
\section*{Acknowledgments}
This work has received funding from the European Research Council (ERC), under the European Union's Horizon 2020 research and innovation programme (FASTPARSE, grant agreement No 714150), from ERDF/MICINN-AEI (ANSWER-ASAP, TIN2017-85160-C2-1-R), from Xunta de Galicia (ED431C 2020/11), and from Centro de Investigación de Galicia ``CITIC'', funded by Xunta de Galicia and the European Union (ERDF - Galicia 2014-2020 Program), by grant ED431G 2019/01. 
\bibliography{iwpt2021}
\bibliographystyle{acl_natbib_nourl}
\begin{appendices}
\section{Parsers}\label{app:parsers}
\paragraph{Left-to-right pointer network} (L2R) is a parser which uses a left to right transition-based algorithm that builds a number of attachments equal to the length of a given sentence together with a pointer network which can point to a given position in the sentence for each token \cite{ma2018stack,fernandez2019left}.\footnote{\url{https://github.com/danifg/SyntacticPointer}} It is one of the current top performing parsers. 
We use the implementation as is, except we make moderate alterations to overcome hardcoded filepaths and the like. Otherwise, the only hyperparameter we change is the number of encoder layers and the number of nodes in the encoder and decoder layers.
\begin{figure}[t!]
    \centering
    \begin{dependency}[edge style={depblue!80, thick},label style={fill=depblue!15},edge slant=7]
    \begin{deptext}[column sep=0.38em,ampersand replacement=\^,font=\footnotesize,baseline=(current bounding box.center)]
 The \^ place \^ had \^ an \^ unco' \^ souch \^ aboot \^ it \\
 \_ \^ $<\!\setminus$ \^ $<\!\setminus$ \^ $/$
  \^ $<$ \^ $<\!\setminus\setminus\!>$ \^ $/$ \^ $<\setminus\!>$ \\
    \end{deptext}
    \depedge{2}{1}{\textsc{det}}
    \depedge{3}{2}{\textsc{nsubj}}
    \depedge{6}{4}{\textsc{det}}
    \depedge{6}{5}{\textsc{amod}}
    \depedge{3}{6}{\textsc{obj}}
    \depedge{6}{8}{\textsc{nmod}}
    \depedge{8}{7}{\textsc{case}}
    \deproot[edge unit distance=3.2ex]{3}{\textsc{root}}
    \end{dependency}%
    \caption{The bracketing encoding from \citet{strzyz2019viable}.  Text is an extract from \citet{cocktails}.}
    \label{fig:sl_example}
\end{figure}
\paragraph{Sequence labelling parser} (\textsc{SeqLab}) is a parsing system that first encodes trees as a set of labels, so that a direct one-to-one prediction can be made for each token in a sentence \cite{spoustov2010dependency,li-etal-2018-seq2seq,strzyz2019viable}.\footnote{We use refactored encoding/decoding functions from \url{https://github.com/mstrise/dep2label}.} We use the original bracketing encoding from \citet{strzyz2019viable} as it doesn't require UPOS tags to decode (as the other leading encoding does), it performs closely to a more recent bracketing encoding that covers more non-projectivity \cite{strzyz2020bracketing}, and the latter encoding wasn't publicly available when this work commenced. It casts a tree as series of tags which are made up of left and right brackets and forward and backwards slashes which encode the incoming and outgoing arcs for each respective node. The encoding for each token is based on edges associated with the preceding tokens and the direction of the edges. More formally, the encoding for $w_i$ is given by:
\begin{center}
\small
\begin{tabular}{rl}
\\
$<$ ---  & if $\;\epsilon_{j(i-1)}\in\mathcal{E} \wedge j>i-1$  \\[.1cm]
$\setminus$ --- & $\times k \;|$
$k=\sum\limits_{w_j\in S}$
$\begin{cases} 
1 & \text{if}\;\; j<i\wedge\epsilon_{ij}\in\mathcal{E}\\
0 &\text{otherwise} 
\end{cases}$ \\[1.25em]
$/$ --- & $\times k \;|$
$k=\sum\limits_{w_j\in S}$
$\begin{cases} 
1 & \text{if}\;\; i\!-\!1<j\wedge\epsilon_{(i-1)j}\in\mathcal{E}\\
0 &\text{otherwise} 
\end{cases}$\\[1.25em]
$>$ --- & if $\;\epsilon_{ji} \wedge j<i$ 
\end{tabular}
\end{center}

We use the biaffine implementation described below and edit it to be a simple sequence-labelling system, i.e.\ an embedding layer, followed by a number of BiLSTM layers, and MLPs one for predicting the bracket tags and one for predicting the edge labels. We use the same hyperparameters as used for the biaffine parser. 

\paragraph{Deep biaffine} (\textsc{Biaffine}) is a graph-based parser that creates two representations of each token from the hidden representations from BiLSTMs, hypothesised to be a representation of each token as dependents and as heads \cite{dozat20161}.\footnote{The original repository (\url{https://github.com/zysite/biaffine-parser}) redirects to a larger set of biaffine based parsers, but is largerly the same.} An affine transformation is applied to the head representation and then this and the dependent one are then combined via a second affine transformation (hence biaffine) to give a matrix of scores, which gives a probability distribution for each node representing the probability any other node is that node's head.  A well-formed tree is then enforced using the Chu–Liu/Edmonds' algorithm \cite{chuliu65,edmonds1967optimum}. The edge labels are then predicted based on the predicted edges. We use the standard hyperparameters for this system except where we match them to better correspond to the L2R parser and then only alter the hyperparameters associated with the BiLSTMs. 

\section{Network hyperparameters}\label{app:hyp}
\begin{table}[htbp!]
\footnotesize
    \centering
        \tabcolsep=.25cm  
    \begin{tabular}{l p{1.5em} r}
    \toprule
    \textbf{Hyperparameter} & & \textbf{Value}\\
    \midrule
         Word embedding dimensions& & 100\\
         Character embedding in ($\lnot$ L2R)&& 32 \\
         Character embedding out ($\lnot$ L2R) && 100 \\
         Character dimension (if L2R) && 100 \\
         Embedding dropout&  & 0.33 \\
         Arc MLP dimensions ($\lnot$ \textsc{SeqLab}) &  & 512\\
         Label MLP dimensions ($\lnot$ \textsc{SeqLab})&  & 128\\
         MLP layers&  & 1 \\
         Epochs&  & 200 \\
         Patience&& 10 \\
         Training batch size && 32 \\
         \bottomrule
    \end{tabular}
    \caption{Hyperparameters for all models. L2R uses a CNN char. embedding layer and \textsc{SeqLab} doesn't have a biaffine layer. Other parameters are as in the original (except \textsc{SeqLab} which uses those of \textsc{Biaffine}).}
    \label{tab:experimental_hyperparameters}
\end{table}
\section{Data}\label{app:data}
\begin{table*}[tb!]
\tabcolsep=0.07cm
    \centering
        \small
\begin{tabular}{l ccccr | ccccr | ccccr}
    \toprule
    & \multicolumn{5}{c|}{Training} & \multicolumn{5}{c|}{Development}
    & \multicolumn{5}{c}{Test} \\
    & Sents. & Tokens & Avg. Len. & NP & Chars. 
    & Sents. & Tokens & Avg. Len. & NP & Chars.   
    & Sents. & Tokens & Avg. Len. & NP & Chars. \\
    \midrule
     \textbf{Chinese} & 15K & 408K & 28.2 & 0.0 & 4.9 & 2K & 51K & 28.0 & 0.0 & 4.9 & 2K & 49K & 27.3 & 0.0 & 4.9 \\
     \textbf{Hindi} & 13K & 281K & 22.1 & 2.6 & 11.4 & 2K & 35K & 22.2 & 2.4 & 11.5 & 2K & 35K & 22.2 & 2.4  & 11.3\\
    \textbf{Korean} & 23K & 296K & 13.9 & 4.5 & 8.2 & 2K &25K & 13.2 & 4.7 & 8.6 & 2K & 28K & 13.4 & 4.0 & 8.3\\
    \textbf{Polish} & 18K & 282K & 16.9 & 1.4 & 5.4 & 2K & 35K & 16.7 & 1.5 & 5.4 & 2K & 34K & 16.2 & 1.4 & 5.4\\
    \bottomrule
    \end{tabular}
    \caption{Treebank statistics: number of sentences (Sents.), number of tokens (Tokens), average sentence length (Avg. Len.), percentage for non-projective arcs (NP), average word length (Chars.).}
    \label{tab:tb_stats}
\end{table*}
We use a small sample of treebanks covering languages from 3 different language families and 4 sub-families and which represent different syntactic systems covering analytic, fusional, and agglutinative languages and all are written in different scripts. We offer a brief description of the treebanks used and some of the salient features of their respective languages. The treebanks were chosen to represent varying syntactic features, but also because of their high quality from being either manually annotated or manually corrected. We also chose relatively large treebanks. The statistics for each treebank are shown in Table \ref{tab:tb_stats}.
\paragraph{UD Hindi-HDTB} (Hindi) is a UD treebank for Hindi based on manually annotated news data \cite{palmer2009hindi,bhat2017hindi}. Hindi is a lightly fusional language with some degree of verbal inflection and noun declension but also makes extensive use of postpositions \cite{McGregor-1977}. It is a split-ergative language meaning in certain cases it uses a nominative-accusative structure but in others it uses an ablative-ergative syntax where the subject of an intransitive verb behaves like the object of a transitive one \cite{Comrie-1978}. It also exhibits tripartite behaviour in certain clauses, where the subject of intransitive verbs, the object of transitive verbs, and the subject of transitive verbs all have different case markings \cite{Comrie-1978}. It is a SOV language, but it has a fairly free word order \cite{Snell-and-Weightman-1989}. It is Indo-Iranian and is written in the Devanagari script.

\paragraph{UD Polish-PDB} (Polish) is a UD treebank manually annotated on fiction, non-fiction, and news data \cite{wroblewska2018extended}. Polish is a highly fusional language with a high degree of verbal inflection \cite{feldstein2001concise} and 7 case-markings \cite{wiese2011optimal}. It is a null-subject language \cite{cognola2018null} with a nominal SVO order but has relatively free word order \cite{Siewierska-1993}. Like most Slavic languages it doesn't make use of articles \cite{Bielec-1998} but it does have a complex system of numeral and quantifiers that result in agreement mistmatches \cite{klockmann2012polish}. It is a Balto-Slavic language written in the Latin script.
\paragraph{UD Korean Kaist} (Korean) is a large treebank generated from a constituency treebank which was semi-automatically annotated with manual corrections based on academic, fiction, and news data \cite{choi1994kaist,chun2018building}. Korean is a strongly suffixing agglutinative language \cite{Ramstedt-1968,Sohn-1999}. This results in a large number of cases and a high degree of verbal inflection \cite{Chang-1996,Song-1988,Lee-and-Ramsey-2000}. It is technically a SOV ordered languae but it has a highly flexible word order \cite{Ramstedt-1968,Sohn-1999}. Korean also uses honorifics and speech levels, the former encoding the social relationship between the speaker and the referents in a discussion and the latter the speaker and the person/people being spoken to \cite{brown2015honorifics}. It is a Koreanic language written in the Hangul script.
\paragraph{Chinese Penn Treebank} (Chinese) is large manually annotated treebank for Mandarin based on news data \cite{xue2002building,xue2005penn}. It is an analytic, isolating language with a SVO dominant word order and is a pro-drop language \cite{Li-and-Thompson-1981}. Chinese has no grammatical tense markers so relies on context or temporal expressions, but aspect is expressed via the use of particles \cite{liu2015tense}. Classifiers and measure words must be used when a noun is preceded by a number, a demonstrative pronoun, or certain quantifiers which are particles that appear between these qualifiers and their respective nouns \cite{her2010semantic}. Chinese is said to be a verb stacking language, where more than one verb or verb phrases are stacked together in the same clause, but there is some disagreement if the way verbs are combined actually constitutes verb stacking \cite{Li-and-Thompson-1981,waltraud2008}. It is a Sino-Tibetan language written in simplified Hanzi. We re-split the data because the standard split has tiny development and test sets. The resulting sizes are shown Table \ref{tab:tb_stats}.

\section{Training time}

\begin{figure}[t!]
    \centering
    \includegraphics[width=0.9\linewidth]{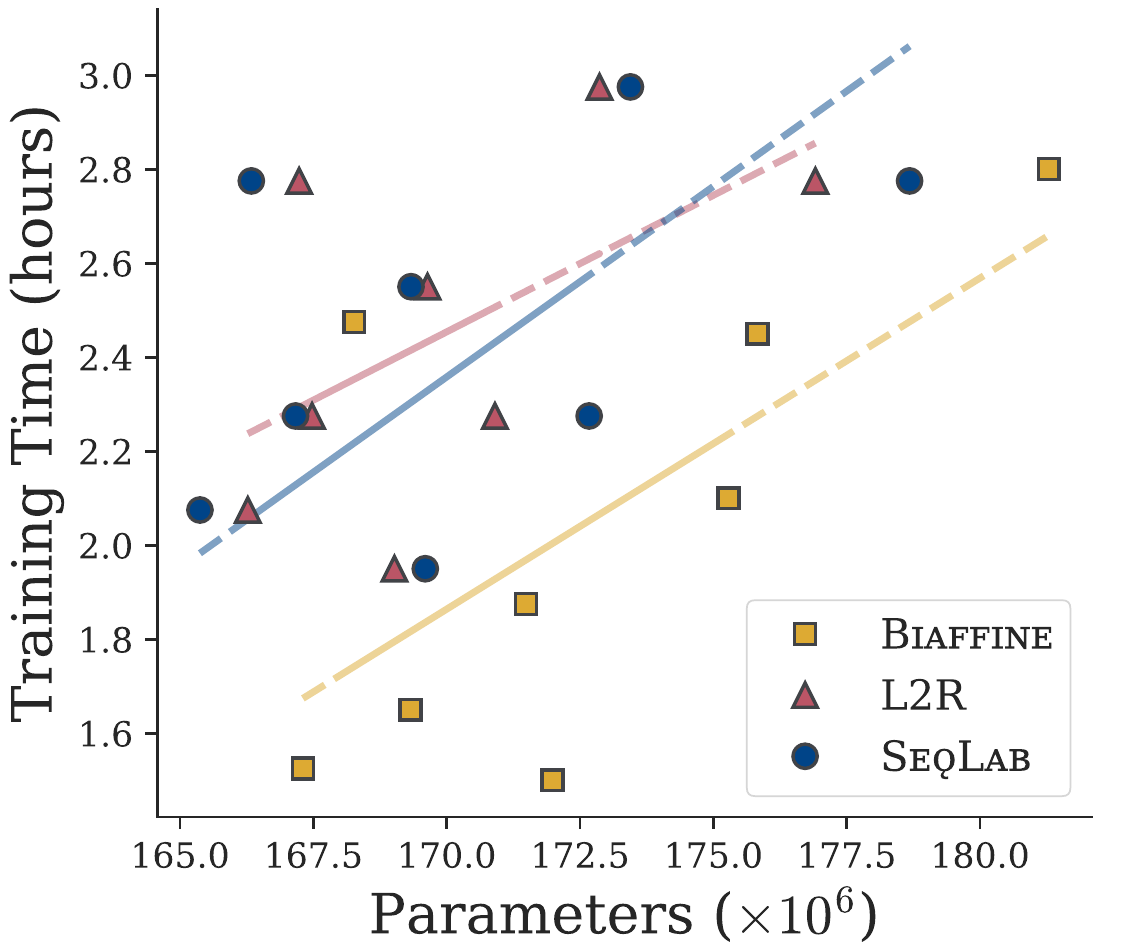}
    \caption{Average training time against total network parameters.}
    \label{fig:time_params}
\end{figure}
\begin{figure}[t!]
    \centering
    \includegraphics[width=0.9\linewidth]{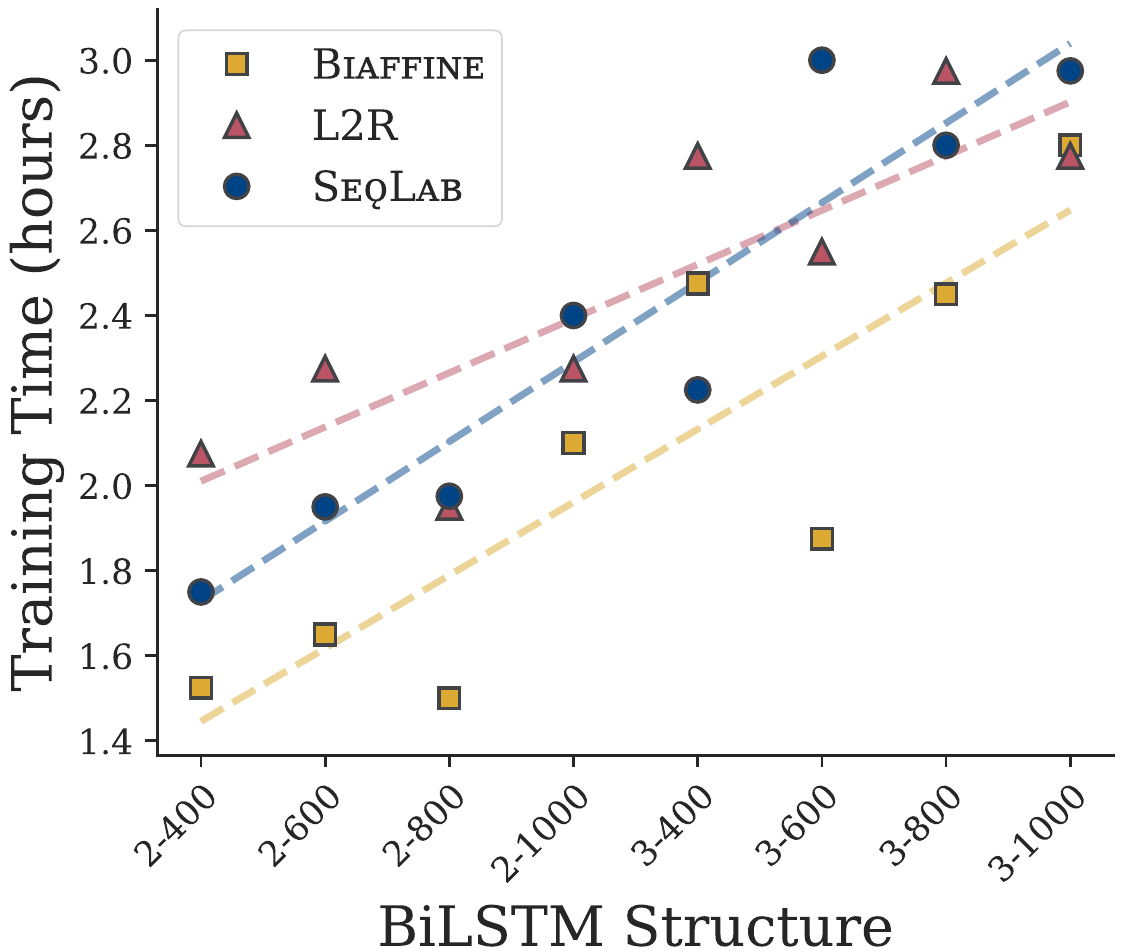}
    \caption{Average training time against BiLSTM structure.}
    \label{fig:time_bilstm}
\end{figure}
\begin{table*}[tb!]
\small
\centering
\tabcolsep=0.075cm
\begin{tabular}{rlccccc p{0.16cm} ccccc p{0.5cm} ccccc}
    \toprule
    \multicolumn{2}{c}{BiLSTM} &  \multicolumn{5}{c}{\textsc{Biaffine}} &&  \multicolumn{5}{c}{\textsc{SeqLab}} &&  \multicolumn{5}{c}{L2R} \\ 
     Layers & Nodes & zh & hi & ko & pl & avg && zh & hi & ko & pl & avg && zh & hi & ko & pl & avg \\
    \midrule
 \multirow{4}{*}{\MyLBrace{5.2ex}{2}} & 400 & 80.59 & 89.83 & 85.35 & 86.22 & 85.50 & & 71.35 & 85.18 & 80.47 & 79.88 & 79.22 & & 79.68 & 89.99 & 83.81 & 85.32 & 84.70 \\ 
      & 600 & 81.33 & 90.48 & 85.43 & 86.88 & 86.03 & & 72.76 & 86.11 & 81.05 & 81.55 & 80.37 & & 80.62 & 90.31 & 84.11 & 86.54 & 85.40 \\
      & 800  & 81.81 & 90.61 & 86.02 & 87.38 & 86.45 & & 74.27 & 86.48 & 81.63 & 82.60 & 81.24 & & 81.59 & 90.03 & 83.97 & 86.77 & 85.59 \\
      & 1000  & 82.15 & 90.56 & 85.91 & 87.95 & 86.64 & & 74.82 & 87.06 & 81.55 & 82.91 & 81.58 & & 81.66 & 90.54 & 84.06 & 87.45 & 85.93 \\[0.5em]
      \multirow{4}{*}{\MyLBrace{5.2ex}{3}} & 400 & 81.71 & 90.55 & 85.45 & 86.83 & 86.13 & & 73.62 & 86.42 & 81.23 & 82.15 & 80.85 & & 82.50 & 90.85 & 84.67 & 87.16 & 86.29 \\
      & 600 & 81.93 & 90.62 & 86.04 & 87.66 & 86.56 & & 75.46 & 87.32 & 81.64 & 83.03 & 81.86 & & 83.25 & 91.13 & 84.75 & 87.83 & 86.74 \\
      & 800 & 82.65 & 91.06 & 85.94 & 88.35 & 87.00 & & 76.20 & 87.65 & 81.66 & 83.90 & 82.35 & & 83.57 & 91.00 & 84.99 & 88.66 & 87.06 \\
      & 1000 & 82.98 & 91.16 & 86.03 & 88.64 & 87.20 & & 76.74 & 87.50 & 81.61 & 84.21 & 82.51 & & 83.41 & 91.20 & 85.28 & 88.84 & 87.18 \\ 
     \bottomrule\\
     \end{tabular}
        \caption{Full LAS results on the development data.}
    \label{tab:full_results_dev}
\end{table*}
\begin{table*}[htbp!]
\small
\centering
\tabcolsep=0.075cm
\begin{tabular}{rlccccc p{0.16cm} ccccc p{0.5cm} ccccc}
    \toprule
    \multicolumn{2}{c}{BiLSTM} &  \multicolumn{5}{c}{\textsc{Biaffine}} &&  \multicolumn{5}{c}{\textsc{SeqLab}} &&  \multicolumn{5}{c}{L2R} \\ 
     Layers & Nodes & zh & hi & ko & pl & avg && zh & hi & ko & pl & avg && zh & hi & ko & pl & avg \\
    \midrule
\multirow{4}{*}{\MyLBrace{5.2ex}{2}} & 400 & 81.03 & 89.74 & 84.58 & 86.76 & 85.53 & & 72.92 & 86.66 & 80.11 & 82.68 & 80.59 & & 79.95 & 90.27 & 83.13 & 85.39 & 84.69 \\
& 600 & 81.82 & 90.39 & 84.89 & 87.38 & 86.12 &  & 74.43 & 87.59 & 80.84 & 83.84 & 81.68 &  & 80.74 & 90.43 & 83.77 & 86.68 & 85.41 \\
& 800 & 82.27 & 90.60 & 85.10 & 88.20 & 86.55 &  & 75.27 & 87.74 & 80.93 & 84.43 & 82.09 &  & 81.73 & 90.27 & 83.57 & 86.76 & 85.58 \\
& 1000 & 82.70 & 90.71 & 84.83 & 88.44 & 86.67 &  & 76.44 & 87.71 & 80.40 & 85.07 & 82.41 &  & 81.86 & 90.35 & 83.73 & 87.42 & 85.84 \\
[0.5em]
      \multirow{4}{*}{\MyLBrace{5.2ex}{3}} & 400  & 81.93 & 90.42 & 84.51 & 87.49 & 86.09 &  & 76.14 & 88.21 & 81.40 & 85.58 & 82.83 &  & 82.63 & 90.98 & 84.57 & 87.59 & 86.44 \\
& 600 & 82.27 & 90.23 & 85.45 & 88.39 & 86.59 &  & 78.13 & 88.77 & 81.96 & 86.69 & 83.89 &  & 83.69 & 91.22 & 84.10 & 88.09 & 86.78 \\
& 800 & 83.11 & 91.08 & 85.46 & 88.78 & 87.11 &  & 78.67 & 88.61 & 81.75 & 86.88 & 83.98 &  & 83.72 & 90.93 & 84.43 & 89.18 & 87.06 \\
& 1000 & 83.47 & 90.94 & 85.56 & 88.86 & 87.21 &  & 78.91 & 89.26 & 81.68 & 87.20 & 84.26 &  & 83.65 & 91.18 & 84.47 & 89.34 & 87.16 \\
        \bottomrule\\
        \end{tabular}
        \caption{Full LAS results on the test data.}
    \label{tab:full_results_test}
\end{table*}

\begin{table*}[t!]
\small
\centering
\tabcolsep=0.075cm
\begin{tabular}{rlccc|ccr| ccc}
    \toprule
    \multicolumn{2}{c}{BiLSTM} &  \multicolumn{3}{c}{Total Energy (MJ)} &  \multicolumn{3}{c}{Total Time (hours)} &\multicolumn{3}{c}{Avg. Parameters ($\times10^6$)}\\
     Layers & Nodes &  \textsc{Biaffine} & \textsc{SeqLab} & \multicolumn{1}{c}{L2R}    & \textsc{Biaffine} & \textsc{SeqLab} & \multicolumn{1}{c}{L2R} & \textsc{Biaffine} & \textsc{SeqLab} & \multicolumn{1}{c}{L2R}   \\
    \midrule
 \multirow{4}{*}{\MyLBrace{5.2ex}{2}} & 400 & 0.54 & 0.59 & 0.69 & 6.1 & 7.0 & 8.3 & 167.3 & 165.4 & 166.3 \\
      & 600 &0.58 & 0.63 & 0.75 & 6.6 & 7.8 & 9.1 & 169.3 & 167.2 & 167.5 \\
      & 800 & 0.53 & 0.68 & 0.65 & 6.0 & 7.9 & 7.8 & 172.0 & 169.6 & 169.0 \\
      & 1000 & 0.73 & 0.83 & 0.76 & 8.4 & 9.6 & 9.1 & 175.3 & 172.7 & 170.9 \\[0.25em]
       \multirow{4}{*}{\MyLBrace{5.2ex}{3}} & 400 &  0.83 & 0.81 & 0.92 & 9.9 & 8.9 & 11.1 & 168.3 & 166.3 & 167.2 \\
      & 600 & 0.66 & 0.96 & 0.84 & 7.5 & 12.0 & 10.2 & 171.5 & 169.3 & 169.6 \\
      & 800 & 0.82 & 0.98 & 1.00 & 9.8 & 11.2 & 11.9 & 175.8 & 173.4 & 172.9 \\
      & 1000 & 1.01 & 1.07 & 0.92 & 11.2 & 11.9 & 11.1 & 181.3 & 178.7 & 176.9 \\
    \bottomrule
\end{tabular}
    \caption{Total energy consumed during training, total training time, and average parameters for each parser system for different BiLSTM configurations.}
    \label{tab:training}
\end{table*}
\begin{figure*}[b!]
    \centering
    \includegraphics[width=0.9\linewidth]{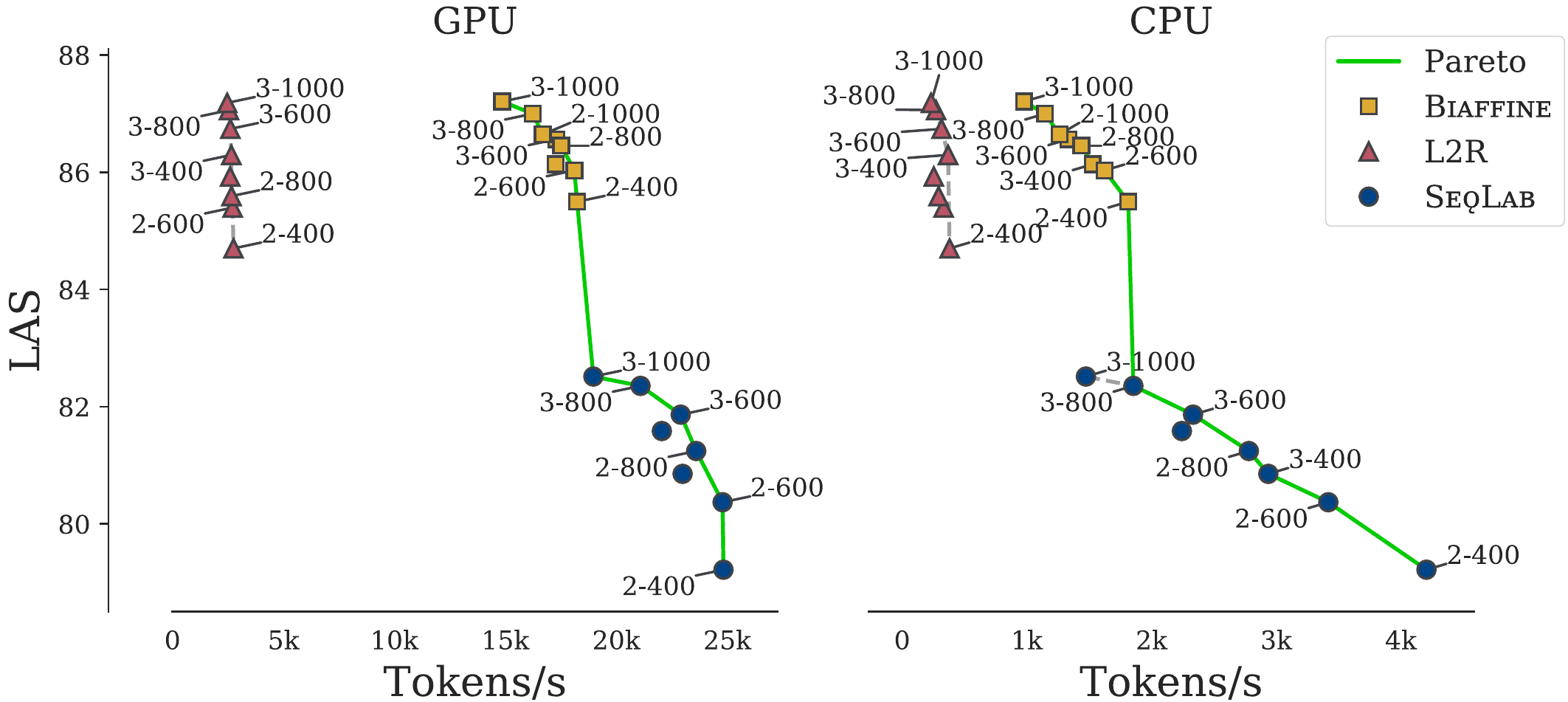}
    \caption{Pareto fronts for L2R, \textsc{Biaffine}, and \textsc{SeqLab} on the test data.}
    \label{fig:speed_paretos_test}
\end{figure*}
Figure \ref{fig:time_bilstm} shows the average training time (across treebanks) for each parser against the BiLSTM structure. There is a clear linear relation as the complexity of the BiLSTM increases. That is considering a BiLSTM with 2 layers and 1000 nodes to be less complex than one with 3 layers and 400 nodes. We also show a similar plot in the Figure \ref{fig:time_params}, but against the total number of parameters in the network, which shows a similar but less clear trend.

\section{Full data}

Table \ref{tab:full_results_dev} shows the full LAS scores for each system for each treebank with different BiLSTM configurations on the development data. Similarly, Table \ref{tab:full_results_test} shows the results for the test data. 
Figure \ref{fig:speed_paretos_test} shows LAS against inference speed for the test data and echoes what was observed for the development data in Figure \ref{fig:speed_paretos}. Table \ref{tab:training} shows the total training energy cost, total training time, and the parameters for each parser and for each BiLSTM configuration.

\section{L2R training efficiency}\label{app:l2r_energy}

\begin{figure}[htbp!]
    \centering
    \includegraphics[width=0.9\linewidth]{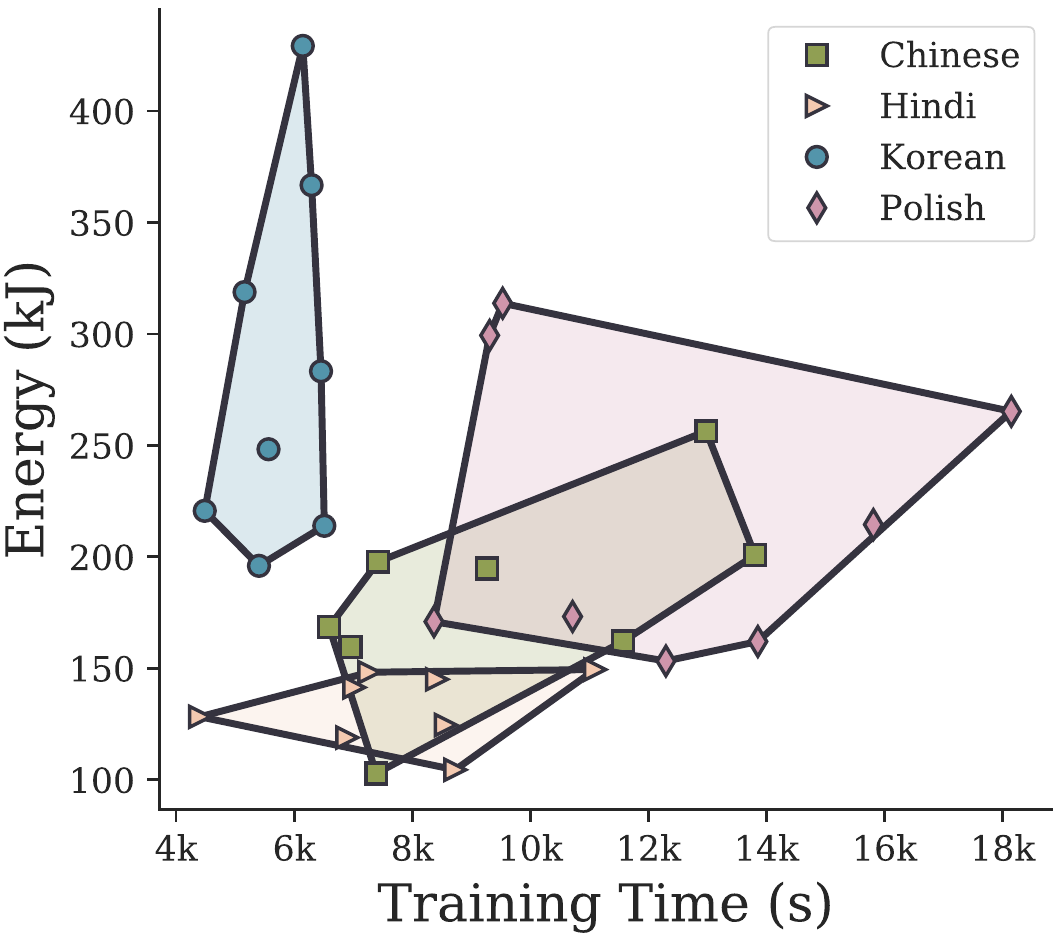}
    \caption{Energy against training time for L2R systems. L2R is more impacted by different data.}
    \label{fig:l2r_energy}
\end{figure}

Figure \ref{fig:l2r_energy} shows training energy against training time for L2R for each treebank used. Clearly, the points associated with each treebank cluster. It is clear training the parser on the Korean data is much more energy consuming compared to the others (which form a linear dispersion). It isn't particularly clear why this would be the case based on the statistics in Table \ref{tab:tb_stats}, except that Korean has the largest number of instances.

\end{appendices}

\end{document}